\documentclass[times, review, 10pt]{elsarticle}

\usepackage{amssymb}
\usepackage[table,xcdraw]{xcolor}
\usepackage{amsmath}
\usepackage{algorithm,algorithmic}
\usepackage{tabularx}
\usepackage{multirow}
\usepackage{booktabs}

\begin{document}

\begin{frontmatter}

\title{Retrieval-Augmented Prompt for OOD Detection}

\author[inst1]{Ruisong Han}
\author[inst1]{Zongbo Han}
\author[inst1]{Jiahao Zhang}
\author[inst2]{Mingyue Cheng}
\author[inst1]{Changqing Zhang}
\address[inst1]{College of Intelligence and Computing, Tianjin University, Tianjin, China}
\address[inst2]{State Key Laboratory of Cognitive
Intelligence, University of Science and Technology of China, China}


\begin{abstract}
Out-of-Distribution (OOD) detection is crucial for the reliable deployment of machine learning models in-the-wild, enabling accurate identification of test samples that differ from the training data distribution. Existing methods rely on auxiliary outlier samples or in-distribution (ID) data to generate outlier information for training, but due to limited outliers and their mismatch with real test OOD samples, they often fail to provide sufficient semantic supervision, leading to suboptimal performance. To address this, we propose a novel OOD detection method called Retrieval-Augmented Prompt (RAP). RAP augments a pre-trained vision-language model's prompts by retrieving external knowledge, offering enhanced semantic supervision for OOD detection. During training, RAP retrieves descriptive words for outliers based on joint similarity with external textual knowledge and uses them to augment the model's OOD prompts. During testing, RAP dynamically updates OOD prompts in real-time based on the encountered OOD samples, enabling the model to rapidly adapt to the test environment. Our extensive experiments demonstrate that RAP achieves state-of-the-art performance on large-scale OOD detection benchmarks. For example, in 1-shot OOD detection on the ImageNet-1k dataset, RAP reduces the average FPR95 by 7.05\% and improves the AUROC by 1.71\% compared to previous methods. Additionally, comprehensive ablation studies validate the effectiveness of each module and the underlying motivations of our approach.
\end{abstract}








\end{frontmatter}


\section{Introduction}
\label{sec:intro}
OOD detection enables models to identify samples that differ from the training distribution during deployment, thereby protecting the model from the negative impacts of OOD data \citep{jia2025enhancing, zhang2025unsupervised}. This capability is especially critical in-the-wild applications, where test samples often differ significantly from the controlled training environment. Such capability is crucial for safety-critical applications like smart healthcare \citep{wei2025fine} and autonomous driving \citep{hu2023planning}, where unreliable decisions due to OOD inputs can lead to serious consequences.

\begin{figure}[t]
    \begin{center}
    \includegraphics[width=\textwidth]{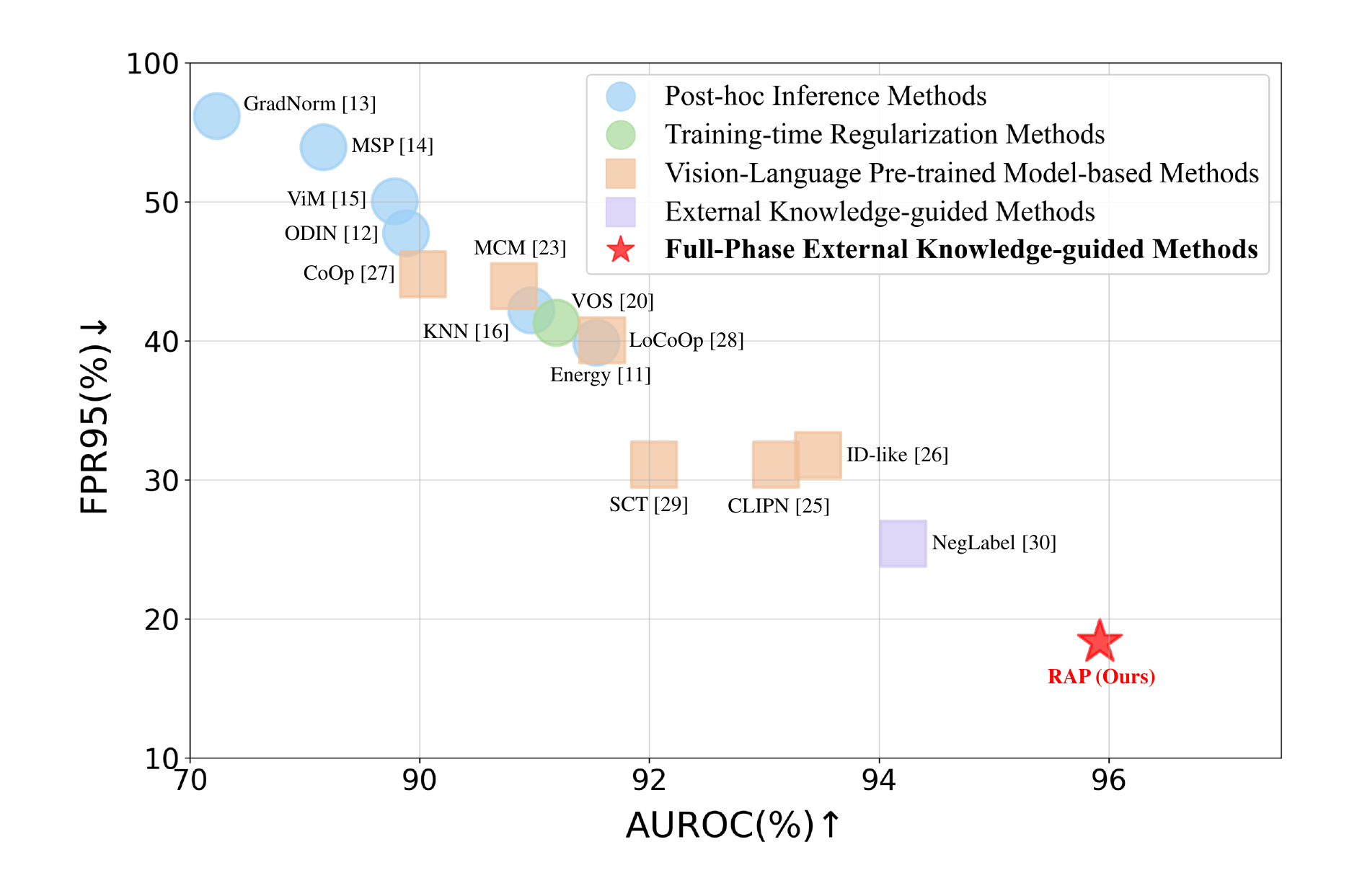}
    \end{center}
    \vspace{-15pt} 
    \caption{
    The evolution of OOD Detection methods.
    }
    \label{evolution}
\end{figure}

As shown in Figure \ref{evolution}, existing OOD detection methods can be broadly categorized into four main types: Post-hoc methods, training-time regularization methods, vision-language pre-trained model-based methods, and external knowledge-guided methods. Post-hoc methods distinguish between in-distribution (ID) and OOD samples by designing various scoring functions. Although convenient, these methods are limited in performance improvement due to the lack of OOD information that can supervise the model. Therefore, some training-time regularization methods explore the use of auxiliary outliers to assist in model training. These methods leverage auxiliary outliers to sample or mine valuable outliers and design various loss functions for learning, thereby enhancing the model's ability to detect OOD samples. However, these methods also face the drawback of being ineffective in scenarios where sufficiently diverse outlier datasets are difficult to obtain. Other studies, based solely on ID data, model the distribution of ID samples and synthesize outliers for learning, which to some extent addresses the issue of unavailable auxiliary outliers. Recently, vision-language models pre-trained on large-scale datasets have shown great potential in various downstream tasks. OOD detection methods based on these models, under zero-shot or few-shot settings, can match or even exceed the performance of original methods that use the entire ID training set or auxiliary outliers, by measuring the similarity between images and ID/OOD prompts. External knowledge can enhance model performance by providing additional context and factual information beyond the training data \citep{cheng2025survey}. To further unlock the potential of vision-language pre-trained models in OOD detection tasks, some methods, for example NegLabel \citep{jiang2024negative}, explore the use of external knowledge. By leveraging the similarity between external knowledge and ID prompts to search for negative prompts, these methods have achieved good detection performance under zero-shot settings. Despite these varied approaches, common challenges remain regarding the effective use of supervision signals and handling distributional shifts.

In summary, many of the methods mentioned above obtain supervision information by utilizing ID samples or auxiliary outliers, and then learn from these samples to enhance the model's ability to detect OOD samples. Although these methods have identified outlier supervision information, there are still some issues with the strategies for learning the OOD detector: 1. Due to the scarcity of valuable outlier samples, previous gradient-based learning methods struggle to effectively utilize these samples for learning, as their limited quantity often fails to provide sufficient supervision signals. 2. There is a distribution difference between the outlier samples used to train the model and the OOD samples encountered in the real test environment, which may further cause the obtained OOD detector to perform poorly. This suggests that relying solely on limited and potentially biased outlier data for gradient-based learning may be insufficient for training robust OOD detectors, motivating the exploration of more stable and richer sources of supervision.

To address the above issues, we leverage external knowledge from large corpora like WordNet \citep{wordnet}, which contains highly refined and complete information. By properly utilizing this knowledge, we can introduce additional semantic supervision. Compared to gradient-based learning methods that use limited outlier supervision, incorporating external semantic supervision for prompt retrieval augmentation leads to better OOD detection performance. We propose a framework for OOD detection based on retrieval-augmented prompts. By leveraging semantic information from knowledge bases and images for retrieval, we can efficiently obtain OOD prompts to facilitate OOD detection. During the training phase, we first use a small number of ID training samples. By randomly cropping training images and computing the similarity between image patches and ID prompts, we obtain valuable outlier image representations. We then retrieve suitable text from external knowledge as OOD prompts based on joint similarity. The joint similarity maximization criterion enables OOD prompts to match the aforementioned valuable outlier image representations while simultaneously distancing from ID representations in both fine-grained image and abstract textual semantics, thereby improving the model's detection performance. During the testing phase, to mitigate the impact of discrepancies between constructed outliers and OOD samples encountered in real testing environments, we continuously adapt to test samples by dynamically perceiving their characteristics and refining OOD prompts. We define our method as the Full-Phase External Knowledge-guided Method. The contributions are summarized as follows:
\begin{itemize}
    \item We propose a framework for OOD detection based on retrieval-augmented prompts, which utilizes the semantic information from external knowledge to augment the model's OOD prompts.
    \item During the testing phase, we utilize potential OOD samples encountered in the testing environment to address the performance degradation caused by discrepancies between outliers constructed during training and real OOD samples.
    \item Through extensive experiments on multiple OOD detection benchmarks, our method achieves state-of-the-art performance. We also conduct in-depth analyses to evaluate the efficiency of our method. Ablation studies on various benchmarks also confirm the effectiveness of retrieval-augmented prompt during both the training and testing phases and the effectiveness of joint similarity.
\end{itemize}

\section{Related Work}
Since the problem of OOD detection was proposed, researchers have introduced various algorithms from different perspectives, including designing different OOD score functions, incorporating auxiliary outlier supervision, utilizing vision-language models, and introducing external knowledge. Specifically, these methods can be categorized into post-hoc methods, training-time regularization methods, vision-language model-based methods, and external knowledge-guided methods. In this section, we provide a detailed introduction to these four categories.

\subsection{Classical OOD Detection Methods}
Classical OOD detection methods mainly refer to approaches that do not rely on vision-language models and are trained or fine-tuned using the entire ID training dataset. These can be further divided into post-hoc methods and training-time regularization methods. 

Post-hoc methods aim to distinguish between ID and OOD data by designing various OOD score functions, requiring no additional training beyond the standard classification model training. Because they do not modify the training process or objective, they are relatively simple and convenient. These methods are particularly valuable in domains with high training costs since retraining is unnecessary. One of the earliest methods is Maximum Softmax Probability (MSP) \citep{DBLP:conf/iclr/HendrycksG17}, which uses the maximum softmax score as the OOD score, serving as an early baseline. ODIN \citep{DBLP:conf/iclr/LiangLS18} enhances OOD detection by applying perturbations to the input and employing temperature scaling on logits, increasing the separability of softmax scores between ID and OOD data. Energy-based OOD detection \citep{DBLP:conf/nips/LiuWOL20} uses the energy score derived from the theoretical relationship between sample energy and its likelihood under the training distribution, improving OOD detection performance. GradNorm \citep{huang2021importance} argues that score functions based solely on output space or feature space often overlook information from the gradient space. It proposes using the gradient norm obtained through backpropagation based on the Kullback-Leibler (KL) divergence between the softmax output and a uniform distribution as the OOD score. KNN-based OOD detection \citep{sun2022knnood} employs a non-parametric approach using the nearest neighbor distance between test data representations and training data representations as the OOD score, overcoming limitations of previous works relying on Gaussian distribution assumptions. ASH \citep{djurisicextremely} improves the separability of OOD scores by truncating and rescaling model output activations. ISH \citep{xuscaling} further explains and analyzes the effective mechanism of ASH and proposes a re-weighting approach during training to enhance OOD detection performance.

Training-time regularization methods can be classified into two categories based on whether they utilize auxiliary outlier data. The first category uses auxiliary outlier datasets and extracts valuable outlier representations as supervision signals. By designing various loss functions, these methods regularize model parameters to enhance OOD detection performance. ATOM \citep{chen2021atom} notes that most auxiliary outliers may not help or may even harm the learning of the OOD detector's decision boundary. It addresses this by identifying and learning from more challenging and informative outliers, resulting in a stronger OOD detector. DivOE \citep{zhu2023diversified} points out that existing auxiliary outlier datasets may lack sufficient feature diversity. It proposes adversarially attacking the outliers to generate noise-augmented outliers with greater feature diversity, thereby enhancing OOD detection. DOS \citep{jiangdiverse} argues that selecting outliers solely based on uncertainty may prevent the model from capturing a comprehensive view of the outlier distribution. To address this, it introduces a simple yet effective method that combines both uncertainty and diversity to identify informative and diverse outliers for model regularization. DiverseMix \citep{yaoout} further increases the diversity of outlier representations by applying Mixup \citep{zhang2018mixup} to outlier data. Theoretically, it demonstrates that greater diversity in auxiliary outliers benefits the learning of OOD detectors. While methods using auxiliary outliers are effective, they often face challenges due to the high cost or unavailability of outlier datasets. To mitigate this, some researchers have explored generating virtual outliers using only the ID training set. VOS \citep{du2022unknown} models the training data representations using a multivariate Gaussian distribution and simulates outliers by sampling from low-likelihood regions. NPOS \citep{tao2023non} improves upon VOS by eliminating the Gaussian distribution assumption and employing a non-parametric approach to simulate outliers, further enhancing the model's OOD detection capabilities. MUFAS \citep{wei2025fine} uses synthetic adversarial OOD samples combined with uncertainty estimation to train an OOD detector for fine-grained medical image OOD detection. MPCL \citep{jia2025enhancing} proposes using diversified multi-prototype contrastive learning during the training phase to enhance the model's OOD detection capability.

\subsection{OOD Detection Methods Based on Visual-Language Pretrained Models}
Vision-language model-based methods can be divided into two categories based on whether they incorporate external knowledge. For the first category, ZOC \citep{esmaeilpour2022zero} uses the pretrained CLIP \citep{DBLP:conf/icml/RadfordKHRGASAM21} model to generate a candidate unknown class name for each test sample. By combining candidate class names with known class names, it calculates a confidence score to achieve zero-shot OOD detection. MCM \citep{mingdelving} proposes a simple and effective zero-shot OOD detection method that leverages the alignment between visual features and textual concepts using CLIP. It uses the maximum classification probability output from CLIP as the OOD score, excelling in complex OOD tasks with semantically similar classes. CLIPN \citep{wang2023clipn} introduces learnable negative prompts and a text encoder to leverage negative semantics from images, significantly improving OOD detection performance. CoOp \citep{zhou2022learning} addresses the downstream adaptation problem of CLIP by proposing a simple and effective prompt learning method. Specifically, CoOp models the context words in prompts as learnable vectors while keeping all pretrained model parameters fixed, achieving automatic prompt optimization. It offers two approaches: unified context and class-specific context, improving the adaptability of vision-language models in downstream tasks, including OOD detection. LoCoOp \citep{DBLP:journals/corr/abs-2306-01293} focuses on few-shot OOD detection using vision-language prompt learning. Unlike traditional prompt learning methods like CoOp, which may suffer from irrelevant information in text embeddings, LoCoOp applies OOD regularization using local features from CLIP. By treating these local features as OOD features and learning to separate them from ID class text embeddings, LoCoOp removes irrelevant interference, enhancing OOD detection capability. ID-like \citep{bai2024id} generates near-ID samples by applying random cropping to few-shot ID datasets. It introduces prompt diversity and prompt alignment regularization to improve CLIP's OOD detection performance. However, while ID-like captures valuable outlier representations, the insufficient outlier supervision signals and the distributional differences between these representations and real test OOD samples reduce the effectiveness of prompt learning. 

In the second category, NegLabel \citep{jiang2024negative} observes that most existing OOD detection algorithms rarely exploit additional information from the text modality. Under a zero-shot setting, NegLabel retrieves negative prompts from the WordNet \citep{wordnet} corpus using cosine similarity with existing ID prompts, further enhancing the OOD detection capabilities of vision-language models. However, NegLabel's introduced external knowledge is often insufficiently accurate, leading to a mismatch between the retrieved knowledge and the true OOD samples, and it cannot quickly adapt to the real OOD samples in the test environment. Unlike previous methods, our proposed approach uses a training-phase OOD prompt retrieval strategy based on the joint similarity maximization principle and a test-phase dynamic OOD prompt retrieval update strategy, introducing more accurate external knowledge. This reduces the gap between the  model's OOD prompts and real test OOD samples, further enhancing the model's OOD detection capability.

\section{Method}
This section first introduces the overall framework of the model, then describes the algorithmic process of RAP during the training and testing phases, and finally explains the model's inference process.

\subsection{Model Framework}
Before introducing the overall model framework, we first describe the problem setting addressed by the proposed method. The proposed RAP is based on a train-time few-shot prompt learning and test-time prompt learning setting. The train dataset $\mathcal{D}= \{x_1,x_2,...x_N\}$ contains only a few samples per class (in the subsequent experiments, we use settings with one sample per class and four samples per class), where $N$ is the total number of training samples, and no auxiliary OOD data is used. Additionally, the algorithm updates prompts in real-time using the test dataset to adapt to the test environment. Next, we introduce the overall framework of RAP.

\begin{figure}[t]
    \begin{center}
    \includegraphics[width=\textwidth]{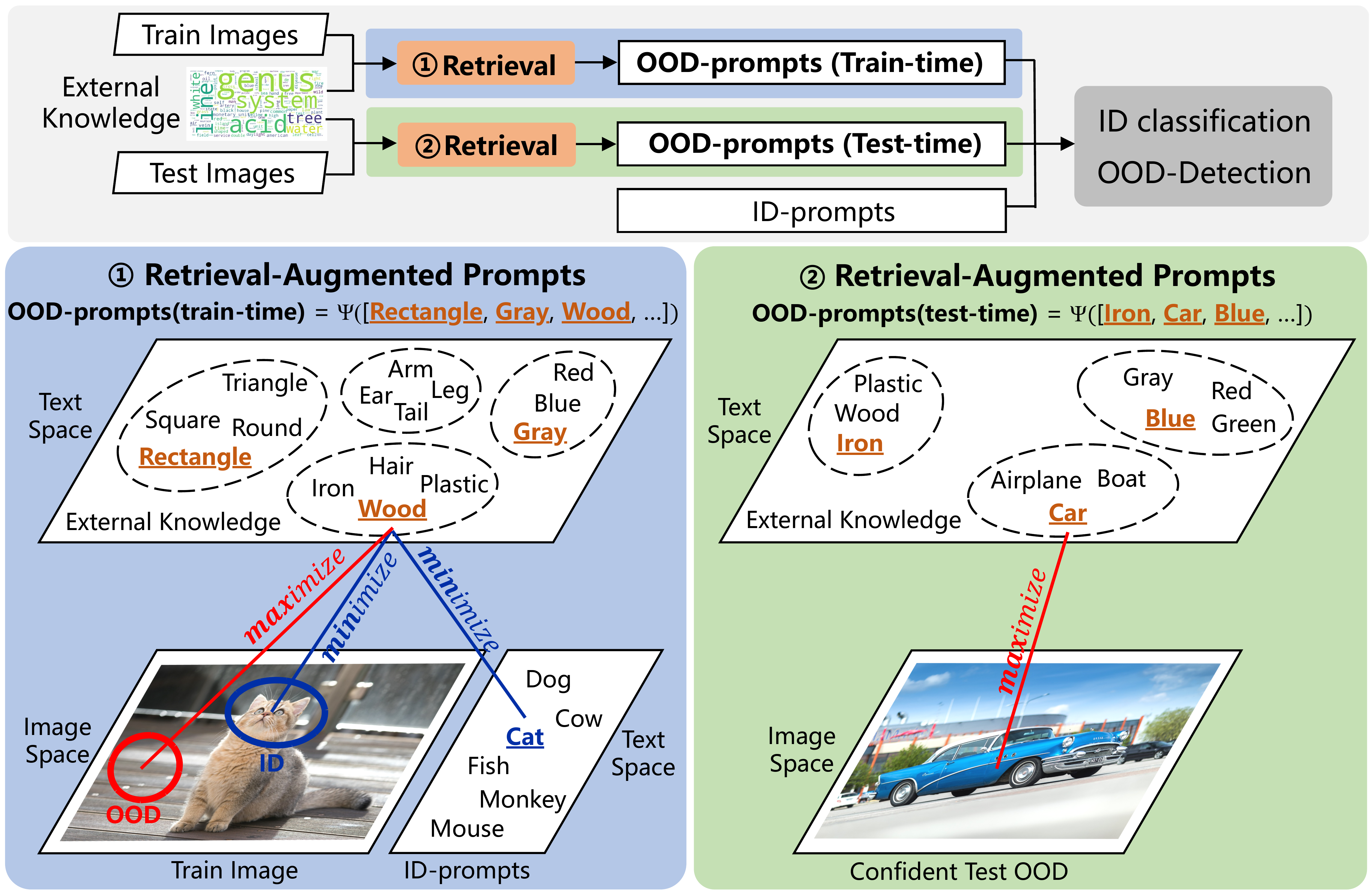}
    \end{center}
    \vspace{-15pt} 
    \caption{
    Overview of the proposed Retrieval-Augmented Prompt (RAP) for Out-of-Distribution Detection. In the training phase (blue part of Fig \ref{fig_framework}), suitable OOD prompts are retrieved based on the the joint similarity between word representations and ID, OOD visual representations, and ID prompts representations, while in the testing phase (green part of Fig \ref{fig_framework}), OOD prompts are retrieved based on the similarity between word representations and confident test OOD representations to adapt to the changing testing environment. ID classification and OOD detection are then performed using ID and OOD prompts.
    }
    \label{fig_framework}
    \vspace{-15pt} 
    
\end{figure}

As shown in Figure \ref{fig_framework}, the proposed method consists of ID prompts $T^{ID}= \{t_{1}^{ID},t_{2}^{ID},...t_{K}^{ID}\}$, OOD prompts $T^{OOD}= \{t_{1}^{OOD},t_{2}^{OOD},...t_{C}^{OOD}\}$, and the visual encoder $\mathcal{I}: x \rightarrow \mathbb{R} ^{d}$ and text encoder $\mathcal{T}: t \rightarrow \mathbb{R} ^{d}$ of the CLIP model \citep{DBLP:conf/icml/RadfordKHRGASAM21}. Classification of ID samples and OOD detection tasks are performed based on the cosine similarity between the ID/OOD prompts and test images. Here, $K$ denotes the number of ID classes, and the OOD prompts $T^{OOD}= \{t_{1}^{OOD},t_{2}^{OOD},...t_{C}^{OOD}\}$ consist of two components: OOD prompts generated during training $T^{OOD}_{train} = \{ t_{1}^{OOD}, t_{2}^{OOD}, \dots, t_{C_{train}}^{OOD} \}$ and OOD prompts generated during testing $T^{OOD}_{test} = \{ t_{1}^{OOD}, t_{2}^{OOD}, \dots, t_{C_{test}}^{OOD} \}$. $C_{train}$ and $C_{test}$ represent the number of OOD prompts generated in the training and testing phases, respectively, with $C$ as the sum of both. These prompts are obtained by retrieving appropriate text from external knowledge to achieve accurate semantic supervision signals. During training (indicated in blue in Figure \ref{fig_framework}), words are retrieved based on their similarity with ID and OOD visual representations as well as ID text prompt representations, which together form a joint similarity. These retrieved words serve as semantic supervision signals to augment OOD prompts during training. During testing (indicated in green in Figure \ref{fig_framework}), words are continuously retrieved and added to the OOD prompts based on their similarity to confidently detected OOD samples. Since the proposed method only enhances OOD prompts, the accuracy of ID classification remains unaffected. The detailed algorithm is described in Algorithm \ref{algorithm:training} and Algorithm \ref{algorithm:testing}.

\subsection{Retrieval-Augmented Prompts During Training}
OOD detection methods typically require high-quality outlier supervision signals obtained from ID data or auxiliary OOD datasets to facilitate subsequent gradient-based prompt learning. Thus, this work constructs valuable outlier representations only from the limited ID training data. Furthermore, due to the difficulty of outlier representations in providing sufficient and effective supervision signals, we use joint similarity to retrieve semantic supervision signals from external knowledge. This retrieved semantic supervision signals is then used as OOD prompts to achieve more effective OOD detection. As shown in Algorithm \ref{algorithm:training}, the procedure consists of two parts: (1) constructing valuable outlier representations from ID training data, and (2) retrieval-augmented prompts during training.

(1) Mining and learning difficult OOD representations can enable the model to more accurately detect OOD samples \citep{ming2022poem,zhu2023diversified,jiangdiverse,du2022unknown,tao2023non,du2024dream, bai2024id}, as these challenging OOD representations often maintain high correlation with ID and frequently occur in the vicinity of ID samples, visually exhibiting characteristics of ``looks like ID but not" \citep{bai2024id}. We hypothesizes that locally sampling ID training images can yield both valuable ID samples and ``near-ID" samples (i.e. outliers) that are insufficiently characterized to be classified as ID. Therefore, we randomly crop a small portion of ID training images and obtain valuable outlier and ID representations based on the cosine similarity between the cropped image patches and ID text prompts \citep{bai2024id}. Since the CLIP model \citep{DBLP:conf/icml/RadfordKHRGASAM21} has strong zero-shot classification capability, we directly use its pre-defined ID prompts and pre-trained visual and text encoders to obtain valuable OOD representations.

Specifically, we first obtain multiple cropped visual representations of the training images. Each ID image $x_i$ is randomly cropped $M$ times to generate $\mathcal{X}^{crop}=\{x_{1,1}^{\text{crop}}, x_{1,2}^{\text{crop}}, \dots, x_{N,M}^{\text{crop}}\}$, and its representations are extracted using the CLIP visual encoder $\mathcal{I}: x \rightarrow \mathbb{R} ^{d}$ to produce $\mathcal{Z}^{crop}=\{z_{1,1}^{\text{crop}}, z_{1,2}^{\text{crop}}, \dots, z_{N,M}^{\text{crop}}\}$.

Next, we select valuable ID and outlier representations using ID prompts and their corresponding cosine similarities. The ID prompts $T^{ID}$ are generated using templates based on the ID class labels. For example, given the label ``dog", the template converts it to ``a photo of a dog", creating a relevant text description. This approach allows the CLIP model to align images and texts for zero-shot classification. The ID prompts are represented as follows:

\begin{equation}
\label{eq:eq3_1}
T^{ID} = \left\{ \text{``a photo of a dog"}, \text{``a picture of a cat"}, \dots \right\}.
\end{equation}

Using the ID prompts $T^{ID}= \{t_{1}^{ID},t_{2}^{ID},...t_{K}^{ID}\}$ and visual representations $z$, the cosine similarity is calculated as follows:

\begin{equation}
\label{eq:eq3_2}
\text{sim}(z, T^{ID}) = \frac{z \cdot T^{ID}}{\|z\| \times \|T^{ID}\|}.
\end{equation}

We select the top $L$ visual patches with the highest and lowest cosine similarity to the ID prompts as the valuable ID representations $\mathcal{Z}^{ID}_{train}$ and the valuable outlier representations $\mathcal{Z}^{OOD}_{train}$, respectively. Formally, for each image's cropped representations $\mathcal{Z}^{crop}_j$, the selection is defined as follows:

\begin{equation}
\label{eq:eq3_3}
\mathcal{Z}^{ID}_{train,j} = \text{topk}(\text{sim}(\mathcal{Z}^{crop}_j, t^{ID}), \mathcal{Z}^{crop}_j, L),
\end{equation}

\begin{equation}
\label{eq:eq3_4}
\mathcal{Z}^{OOD}_{train,j} = \text{topk}(-\text{sim}(\mathcal{Z}^{crop}_j, t^{ID}), \mathcal{Z}^{crop}_j, L),
\end{equation}

where $\text{sim}(\mathcal{Z}^{crop}_j, t^{ID})$ represents the set of similarities, and $\text{topk}(A, B, L)$ denotes selecting the top $L$ elements from set $B$ based on the corresponding values in set $A$. The selected valuable ID and outlier representations are subsequently used for retrieval-augmented prompt learning during the training phase.

(2) Existing gradient-based prompt learning OOD detection methods face the challenge of insufficient supervision due to the scarcity of valuable outlier samples in few-shot settings. This problem makes it difficult for these methods to provide sufficiently accurate and effective supervision signals, limiting their OOD detection capability. To address this issue, as illustrated in the blue part of Figure \ref{fig_framework}, RAP retrieves words from the external large-scale structured text corpus WordNet \citep{wordnet} as additional semantic supervision. These external semantic supervision signals are directly used as OOD prompts during the training phase for subsequent OOD detection, called Retrieval-Augmented Prompts(Train-time). These words are selected to match valuable outlier representations while avoiding overlap with ID representations in both abstract fine-grained image and text semantic levels to ensure accuracy and effectiveness.

To achieve this goal, our method applies a joint similarity maximization principle during the retrieval process. Specifically, for all nouns and adjectives in WordNet \citep{wordnet}, we use CLIP \citep{DBLP:conf/icml/RadfordKHRGASAM21}'s text encoder to extract word representations $H^{W}=\{ h^{W}_i \}_{i=1}^{N_W}$ and calculate the joint similarity $sim_{train}$ with $\mathcal{Z}^{OOD}_{train}$, $\mathcal{Z}^{ID}_{train}$, and $h^{ID}$ as follows:

\begin{equation}
\label{eq:eq3_5}
    sim_{train} = \lambda_{1}sim_{1}+\lambda_{2}sim_2+\lambda_{3}sim_3 \text{,}
\end{equation}

where $sim_{1}$, $sim_{2}$, and $sim_{3}$ represent the similarity between the word representation $h^{W}_i$ and $\mathcal{Z}^{OOD}_{train}$, $h^{W}_i$ and $\mathcal{Z}^{ID}_{train}$, and $h^{W}_i$ and $h^{ID}$, respectively. $\lambda_{1}$, $\lambda_{2}$, and $\lambda_{3}$ are coefficients balancing these similarities ($\lambda_{1}>0$, $\lambda_{2}<0$, $\lambda_{3}<0$). To obtain the OOD prompts, we first select the top $P$ words with the highest joint similarity scores. Specifically, the selected nouns and adjectives are incorporated into prompt templates $\Psi(\cdot)$ such as ``the nice \textless n.\textgreater'' and ``This is a \textless adj.\textgreater\ photo'' to construct the OOD prompts used during training, formulated as:

\begin{equation}
\label{eq:eq3_6}
T^{OOD}_{train} = \Psi(\text{topk}(sim_{train}, H^{W}, P)) \text{.}
\end{equation}

Maximizing $sim_{1}$ helps obtain accurate external knowledge with semantic supervision aligned with outliers. This alleviates the lack of supervision signals caused by the scarcity of valuable outlier representations. Meanwhile, minimizing $sim_{2}$ and $sim_{3}$ ensures that the retrieved external knowledge remains distant from ID sample representations in both fine-grained image and abstract text semantics, reducing semantic overlap between ID and OOD prompts. Specifically, fine-grained image semantics capture detailed visual information such as object shapes, textures, and colors, which help differentiate OOD and ID samples. On the other hand, abstract text semantics focus on high-level concepts, including object categories and properties, supporting broader sample classification. By comparing ID representations with external knowledge at both levels, we mitigate semantic ambiguities and improve OOD detection accuracy and robustness.

To further prevent local optima, we compute the average similarity between each word representation $h^{W}_{i}$ and all ID/outlier visual representations $z$ for $sim_{1}$ and $sim_{2}$ as follows:

\begin{equation}
\label{eq:eq3_7}
    sim_1=\frac{1}{N\times S} \sum\nolimits_{j=1}^{N} \sum\nolimits_{l=1}^{L}sim(h^{W}_{i},z_{j,l}^{OOD}) \text{,}
\end{equation}

\begin{equation}
\label{eq:eq3_8}
    sim_2=\frac{1}{N\times S} \sum\nolimits_{j=1}^{N} \sum\nolimits_{l=1}^{L}sim(h^{W}_{i},z_{j,l}^{ID}) \text{.}
\end{equation}

To enhance retrieval robustness \citep{jiang2024negative}, we calculate $sim_{3}$ using the $\eta$-percentile similarity between word representations and ID prompt representations, as follows:

\begin{equation}
\label{eq:eq3_9}
    sim_3= \text{percentile}_{\eta}(  \{ sim(h^{W}_i,h^{ID}_{j})   \}_{j=1}^{K}) \text{.}
\end{equation}
\begin{algorithm}[t]
    \caption{Retrieval-Augmented Prompts During Training}
    \label{algorithm:training}
    \renewcommand{\algorithmicrequire}{\textbf{Input:}}
    \renewcommand{\algorithmicensure}{\textbf{Output:}}
    \begin{algorithmic}[1]
        \STATE \textbf{Input:} Training data: $\mathcal{D}= \{x_1,x_2,...x_N\}$.
        \STATE \textbf{Initialization:} Pretrained CLIP model with image encoder $\mathcal{I}: x \rightarrow \mathbb{R}^{d}$ and text encoder $\mathcal{T}: t \rightarrow \mathbb{R}^{d}$.
        
        \STATE \textbf{/* Constructing Valuable Outlier representations from ID training data */}
        \FOR{$i = 1, 2, \ldots, N$}
            \STATE Perform random cropping on each image $x_i$ to obtain $\mathcal{X}^{crop}$.
            \STATE Use $\mathcal{I}$ to extract visual representations $\mathcal{Z}^{crop}$ from $\mathcal{X}^{crop}$.
            \STATE Generate ID prompts using Equation \ref{eq:eq3_1}.
            \STATE Obtain valuable ID representations using Equation \ref{eq:eq3_3}.
            \STATE Obtain valuable outlier representations using Equation \ref{eq:eq3_4}.
        \ENDFOR
        
        \STATE Collect all representations as $\mathcal{Z}^{ID}_{train}$ and $\mathcal{Z}^{OOD}_{train}$.
        \STATE Compute $sim_1$, $sim_2$, $sim_3$ using Equations \ref{eq:eq3_7}, \ref{eq:eq3_8}, \ref{eq:eq3_9}.
        \STATE Calculate $sim_{train}$ using Equation \ref{eq:eq3_5}.
        
        \STATE \textbf{/* RAP During Training */}
        \STATE Retrieve and augment OOD prompts from WordNet using Equation \ref{eq:eq3_6}.
        
        \STATE \textbf{Output:} ID prompts $T^{ID}$ and OOD prompts $T^{OOD}_{train}$.
    \end{algorithmic}  
\end{algorithm}

\subsection{Retrieval-Augmented Prompts During Testing}

Most existing OOD detection methods optimize prompts only during the training phase, making them challenging to adapt quickly and effectively to test environments. To address this issue, we proposes a method to rapidly update OOD prompts at test time based on test samples. As shown in the green section of Figure \ref{fig_framework}, we first obtains confident and valuable OOD representations based on the ID scores of test samples. Then, it retrieves appropriate words from external knowledge and adds them to the existing OOD prompts to achieve online adaptation to the test environment. Specifically, as shown in Algorithm \ref{algorithm:testing}, the process consists of two main components: (1) acquiring OOD samples from the test data; (2) retrieval-augmented prompts during testing.

(1) During the test phase, since real OOD samples are available, we does not construct OOD image representations from ID test samples. Instead, we select confident and valuable outliers from the test data as supervision signals based on the OOD detection scores predicted by the model. Specifically, test samples with scores $S(z)$ between $u_1$ and $u_2$ are identified as valuable OOD samples. This intermediate score threshold ensures that the model accurately detects these samples, which are also relatively valuable and likely to be true OOD samples, thereby providing more effective supervision signals to the model, as shown in the following equation:

\begin{equation}
\label{eq:eq3_10}
    \mathcal{Z}_j^{testOOD} = \left\{ z_j \mid u_1 \leq S(z_j) \leq u_2 \right\}.
\end{equation}

(2) To adapt to the changing testing environment in time, RAP continuously augments OOD prompts with external knowledge by learning from valuable test OOD samples. Specifically, as illustrated in the green section of Figure \ref{fig_framework}, the similarity $sim_{test}=sim(h^{W}_i,z_j^{testOOD})$ between word representations $h^{W}_i$ and valuable test OOD image representations $z_j^{testOOD}$ is calculated. The top $Q$ words with the highest similarity to the test OOD samples are selected as new OOD prompts $T^{OOD}_{test, \space new}$ and added to the existing test-time OOD prompts $T^{OOD}_{test}$. The selected new OOD prompts $T^{OOD}_{test, \space new}$ are defined by the following equation:

\begin{equation}
\label{eq:eq3_11}
T^{OOD}_{test, \space new} = \Psi(\text{topk}(sim_{test}, H^{W}, Q)).
\end{equation}

The OOD prompts at test time are continuously updated to ensure adaptation to the changing test environment.

\subsection{ID Classification and OOD Detection}

This process consists of two main components: (1) ID classification; (2) OOD detection.

(1) The ID classification results are obtained based on the cosine similarity between the visual representation $z$ and each ID prompt representation $t^{ID}$, as defined by the following equation:

\begin{equation}
\label{eq:eq3_12}
    \hat{y} = \underset{y_i \in \mathcal{Y}}{\arg\max \limits}  \{ sim(z, t_i^{ID})  \}.
\end{equation}

(2) By calculating the cosine similarity between the visual representation and both ID and OOD prompt representations, the ID score $S(x)$ is obtained using the following equation:

\begin{equation}
\label{eq:eq3_13}
    S(x)=\frac{\sum_{i=1}^{K} \varphi(z,h^{ID}_{i})}{ \sum_{i=1}^{K} \varphi(z,h^{ID}_{i}) + \sum_{i=1}^{C} \varphi(z,h^{OOD}_{i})},
\end{equation}

where $\varphi(z,h)=e^{\tfrac{sim(z,h)}{\tau }}$, and $\tau$ is the temperature parameter from CLIP \citep{DBLP:conf/icml/RadfordKHRGASAM21}. The proposed method constructs an OOD detector $F(x)$ to perform the OOD detection task based on the OOD detection score $S(x)$, as described in the following equation:

\begin{equation}
\label{eq:eq3_14}
    F(x)=\begin{cases}
    \text{ID}, & S(x)\ge \gamma \\
    \text{OOD}, & S(x)<\gamma
    \end{cases},
\end{equation}

where $\gamma$ is the threshold to determine whether a sample is ID or OOD.

Increasing OOD prompts during continuous retrieval augmentation may lead to semantic overlap with ID samples \citep{jiang2024negative}, reducing OOD detection performance. To mitigate this issue, the proposed method follows the prompt ensemble strategy used in NegLabel \citep{jiang2024negative}. The OOD prompts are randomly divided into $N_g$ groups, where each group of OOD prompts is combined with all ID prompts to calculate the ID score. The average score across all groups is used as the final OOD detection score, defined as follows:

\begin{equation}
\label{eq:eq3_15}
    S_{g}(x)=\frac{1}{N_g} \sum\nolimits_{j=1}^{N_g}S_j(x).
\end{equation}

\begin{algorithm}
    \caption{Retrieval-Augmented Prompts During Testing}
    \label{algorithm:testing}
    \renewcommand{\algorithmicrequire}{\textbf{Input:}}
    \renewcommand{\algorithmicensure}{\textbf{Output:}}
    \begin{algorithmic}[1]
        \STATE \textbf{Input:} Test data $x_{test}$, trained prompts $T^{ID}$ and $T^{OOD}_{train}$.
        
        \STATE \textbf{/* Obtain Valuable OOD Representations */}
        \STATE Compute OOD detection scores using Equation \ref{eq:eq3_15}.
        \STATE Identify valuable OOD samples using Equation \ref{eq:eq3_10}.
        
        \STATE \textbf{/* RAP During Testing */}
        \IF{valuable OOD samples exist}
            \STATE Generate new OOD prompts $T^{OOD}_{test, \space new}$ using Equation \ref{eq:eq3_11}.
            \STATE Update OOD prompts: $T^{OOD}_{test} \leftarrow T^{OOD}_{test} \cup T^{OOD}_{test, \space new}$.
        \ENDIF
        
        \STATE \textbf{/* Inference */}
        \STATE Perform ID classification using Equation \ref{eq:eq3_12}.
        \STATE Perform OOD detection using Equation \ref{eq:eq3_14}.
        
        \STATE \textbf{Output:} ID predictions and OOD detection results.
    \end{algorithmic}  
\end{algorithm}

\section{Experiments}

This section provides a detailed description of the experimental setup, results, analysis, and various ablation studies.

\subsection{Experimental Design}

The experiments in RAP consist of the following parts:  
(1) Main experiments, where the performance of the proposed method is compared with other baseline methods on two large-scale OOD detection benchmark datasets.  
(2) Effectiveness of prompt retrieval augmentation applied in both the training and testing phases.  
(3) Effectiveness of the joint similarity maximization principle for prompt retrieval augmentation during the training phase.  
(4) Impact of the number of retrieved prompts on detection performance.

\subsection{Dataset Description}

Following the settings of MCM \citep{mingdelving}, NegLabel \citep{jiang2024negative}, and ID-like \citep{bai2024id}, the proposed method is comprehensively evaluated on both standard and challenging OOD detection benchmarks. In the standard benchmark, ImageNet-1k \citep{huang2021importance} is used as the ID dataset, while iNaturalist \citep{DBLP:conf/cvpr/HornASCSSAPB18}, Places \citep{DBLP:journals/pami/ZhouLKO018}, SUN \citep{DBLP:conf/cvpr/XiaoHEOT10}, and Textures \citep{DBLP:conf/cvpr/CimpoiMKMV14} are used as OOD datasets. For the challenging benchmark, ImageNet-10 \citep{mingdelving}, ImageNet-20 \citep{mingdelving}, and ImageNet-100 \citep{mingdelving} are alternately used as ID and OOD datasets, ensuring no semantic overlap between ID and OOD data. The details of these datasets are as follows:

(1) iNaturalist \citep{DBLP:conf/cvpr/HornASCSSAPB18} is a fine-grained dataset containing 859,000 images of over 5,000 species of plants and animals. The maximum edge length of each image is resized to 800 pixels. A total of 110 plant species not present in ImageNet-1k \citep{huang2021importance} are manually selected, and 10,000 images from these categories are randomly sampled as the OOD dataset for the standard benchmark.  

(2) SUN \citep{DBLP:conf/cvpr/XiaoHEOT10} contains 397 categories and over 130,000 images, with all images larger than $200 \times 200$ pixels. Since SUN \citep{DBLP:conf/cvpr/XiaoHEOT10} has some category overlaps with ImageNet-1k, 50 unique nature-related categories (e.g., forests, icebergs) are selected, and 10,000 images are randomly sampled as the OOD dataset for the standard benchmark.  

(3) Places365 \citep{DBLP:journals/pami/ZhouLKO018} is a scene dataset conceptually similar to SUN \citep{DBLP:conf/cvpr/XiaoHEOT10}, with images resized to ensure a minimum dimension of 512 pixels. 50 categories not present in ImageNet-1k are manually selected, and 10,000 images from these categories are used as the OOD dataset for the standard benchmark.  

(4) Textures \citep{DBLP:conf/cvpr/CimpoiMKMV14} contains 5,640 texture images with dimensions ranging from $300 \times 300$ to $640 \times 640$ pixels. The entire dataset is used as the OOD dataset for the standard benchmark.  

(5) ImageNet-1k \citep{huang2021importance} covers a wide variety of real-world object categories. Compared to commonly used CIFAR datasets \citep{krizhevsky2009learning}, ImageNet-1k contains over ten times more labels, and its image resolution is significantly higher than CIFAR ($32 \times 32$) or MNIST ($28 \times 28$). ImageNet-1k is used as the ID dataset for the standard benchmark.  

(6) ImageNet-10, ImageNet-20, and ImageNet-100 \citep{mingdelving} are subsets of ImageNet-1k, with categories exhibiting certain semantic similarities, making them more challenging to distinguish. For example, distinguishing dogs (ID) from wolves (OOD). These datasets are alternately used as ID and OOD datasets in the challenging benchmark.

\subsection{Evaluation Metrics}

The proposed method is evaluated using two commonly used metrics in OOD detection tasks: (1) FPR95: Measures the false positive rate when the true positive rate is 95\%. A lower FPR95 indicates that the model is better at distinguishing between ID and OOD samples. (2) AUROC: Represents the area under the receiver operating characteristic (ROC) curve. A higher AUROC value (closer to 1) indicates better model performance.  

\subsection{Baseline Methods}

The proposed method is compared with various OOD detection methods, including Post-hoc methods (fully supervised), Training-time regularization methods (fully supervised), Vision-language pre-trained model-based methods (zero-shot or few-shot), and External knowledge-guided methods (zero-shot).

Post-hoc methods: MSP \citep{DBLP:conf/iclr/HendrycksG17} uses the maximum softmax output probability as the OOD score. ODIN \citep{DBLP:conf/iclr/LiangLS18} improves MSP by applying input perturbations and temperature scaling. Energy \citep{DBLP:conf/nips/LiuWOL20} computes the OOD score using an energy function. GradNorm \citep{huang2021importance} uses the vector norm of the gradients as the OOD score. ViM \citep{haoqi2022vim} combines feature residual norms and principal spaces to calculate the OOD score. KNN \citep{sun2022knnood} performs non-parametric nearest-neighbor-based OOD detection.

Training-time regularization methods: VOS \citep{du2022unknown} synthesizes virtual OOD samples using a class-conditional Gaussian distribution. NPOS \citep{tao2023non} extends VOS by generating synthetic OOD data without assuming the distribution of ID embeddings.

Vision-language pre-trained model-based methods: Mahalanobis \citep{DBLP:conf/nips/LeeLLS18}and Energy \citep{DBLP:conf/nips/LiuWOL20} were originally used as post-hoc methods for OOD detection, where Mahalanobis distance and Energy function were employed as OOD detection scores. Following NegLabel \citep{jiang2024negative}, we apply them to zero-shot OOD detection in vision-language models. ZOC \citep{esmaeilpour2022zero} expands CLIP \citep{DBLP:conf/icml/RadfordKHRGASAM21} by generating candidate unknown class names. MCM \citep{mingdelving} uses the maximum classification probability from CLIP. CLIPN \citep{wang2023clipn} introduces learnable negative prompts and text encoders for OOD detection. CoOp \citep{zhou2022learning} uses conventional prompt learning for OOD detection. LoCoOp \citep{DBLP:journals/corr/abs-2306-01293} enhances OOD detection by leveraging local features in CLIP. ID-like \citep{bai2024id} generates near-ID samples using random crops and applies prompt learning. SCT \citep{yu2025self} introduces adaptive modulation of the OOD regularization influence to enhance prompt-based OOD detection performance, particularly with limited ID data.

External knowledge-guided methods: NegLabel \citep{jiang2024negative} retrieves negative prompts from WordNet \citep{wordnet} based on cosine similarity to further enhance OOD detection.

\subsection{Parameter Settings}
All methods, including RAP, adopt CLIP/ViT-B/16 as the pre-trained model. For each class in the training set, we randomly select 1 and 4 samples as the few-shot training dataset. During the training phase, when constructing outlier representations from ID training data, each sample undergoes $M$ random crops, and the $L$ samples with the highest similarity to the predefined prompts, as well as the $L$ samples with the lowest similarity, are selected as valuable ID and outlier data, respectively. When $M$ is too large, excessive cropping can lead to increased computational costs and wasted resources. Conversely, when $M$ is too small, it becomes challenging to perform sufficient cropping of the training images to obtain valuable ID and outlier representations. Similarly, if $L$ is too small, the number of valuable representations may be insufficient, negatively impacting OOD detection performance. On the other hand, if $L$ is too large, the selected ID and outlier representations may contain excessive noise and irrelevant information. Therefore, referring to ID-like \citep{bai2024id}, we choose moderate values for $M$ and $L$, setting $M$ to 256 and $L$ to 32 in our experiments.
\begin{table}[h!]
\centering
\caption{Hyper Parameter Settings for Different Datasets}
\label{hyper_params}
\begin{tabular}{ccccccc}
\toprule[1.5pt]
\multirow{2.5}{*}{ID} & \multirow{2.5}{*}{OOD} & \multicolumn{5}{c}{Hyper Parameters} \\ 
\cmidrule(lr){3-7}
& & $\lambda_1$ & $\lambda_2$ & $\lambda_3$ & $u_1$ & $u_2$ \\ 
\midrule

\multirow{4}{*}{ImageNet-1K} & iNaturalist & 0.2 & -0.2 & -1 & 0.5 & 0.6\\ 
& SUN & 0.2 & -0.2 & -1 & 0.2 & 0.3 \\ 
& Places & 0.2 & -0.2 & -1 & 0.4 & 0.5 \\ 
& Textures & 0.2 & -0.2 & -1 & 0.4 & 0.6 \\ 
\midrule
ImageNet-10 & ImageNet-20 & 0.05 & -0.005 & -1 & 0.0 & 0.5 \\ \midrule
ImageNet-20 & ImageNet-10 & 0.1 & -0.02 & -1 & 0.0 & 0.2 \\ \midrule
ImageNet-100 & ImageNet-10 & 0.1 & -0.01 & -1 & 0.0 & 0.2 \\ 

\bottomrule[1.5pt]
\end{tabular}
\end{table}

In the training phase of RAP, the number of prompts for retrieval augmentation $P$ is set to 10,000. The $\eta$-percentile value is set to 5. $\tau$ is set to 0.01. During the test phase, the number of prompts used for retrieval augmentation per test sample $Q$ is set to 4. The number of groups in the prompt ensemble is set to 100. The weight coefficients $\lambda_{1}$, $\lambda_{2}$, and $\lambda_{3}$ in the joint similarity and the values of $u_1$ and $u_2$ are shown in the Table \ref{hyper_params}.

\begin{table}[]
\caption{OOD detection performance comparison with multiple baselines on Standard Benchmark. All values are percentages. ↑ indicates larger values are better and ↓ indicates smaller values are better. Bold \textcolor{red}{\textbf{red}} indicates the best, bold \textcolor{blue}{\textbf{blue}} indicates the second best.}
\label{standardOOD}
\resizebox{1\textwidth}{!}{
\begin{tabular}{c|cccccccc|cc}
\toprule[1.5pt]
                     & \multicolumn{8}{c|}{OOD Dataset}                               & \multicolumn{2}{c}{} \\
 &
  \multicolumn{2}{c}{iNaturalist} &
  \multicolumn{2}{c}{SUN} &
  \multicolumn{2}{c}{Places} &
  \multicolumn{2}{c|}{Textures} &
  \multicolumn{2}{c}{\multirow{-2}{*}{Average}} \\ \cmidrule[1.0pt](l){2-3} \cmidrule[1.0pt](l){4-5}\cmidrule[1.0pt](l){6-7}  \cmidrule[1.0pt](l){8-9}\cmidrule[1.0pt](l){10-11}
\multirow{-3}{*}{Methods} &
  AUROC↑ &
  FPR95↓ &
  AUROC↑ &
  FPR95↓ &
  AUROC↑ &
  FPR95↓ &
  AUROC↑ &
  FPR95↓ &
  AUROC↑ &
  FPR95↓ \\ \midrule[1.0pt]
                     & \multicolumn{10}{c}{Full/Sub Data Fine-tune}                  \\
MSP \citep{DBLP:conf/iclr/HendrycksG17}                & 87.44 & 58.36 & 79.73 & 73.72 & 79.67 & 74.41 & 79.69 & 71.93 & 81.63     & 69.61    \\
ODIN \citep{DBLP:conf/iclr/LiangLS18}               & 94.65 & 30.22 & 87.17 & 54.04 & 85.54 & 55.06 & 87.85 & 51.67 & 88.80     & 47.75    \\
Energy \citep{DBLP:conf/nips/LiuWOL20}            & 95.33 & 26.12 & 92.66 & 35.97 & 91.41 & 39.87 & 86.76 & 57.61 & 91.54     & 39.89    \\
GradNorm \citep{huang2021importance}           & 72.56 & 81.50 & 72.86 & 82.00 & 73.70 & 80.41 & 70.26 & 79.36 & 72.35     & 80.82    \\
ViM \citep{haoqi2022vim}                 & 93.16 & 32.19 & 87.19 & 54.01 & 83.75 & 60.67 & 87.18 & 53.94 & 87.82     & 50.20    \\
KNN \citep{sun2022knnood}               & 94.52 & 29.17 & 92.67 & 35.62 & 91.02 & 39.61 & 85.67 & 64.35 & 90.97     & 42.19    \\
VOS \citep{du2022unknown}            & 94.62 & 28.99 & 92.57 & 36.88 & 91.23 & 38.39 & 86.33 & 61.02 & 91.19     & 41.32    \\
NPOS \citep{tao2023non}            & 96.19 & 16.58 & 90.44 & 43.77 & 89.44 & 45.27 & 88.80 & 46.12 & 91.22     & 37.93    \\ \midrule[1.0pt]
\multicolumn{1}{l}{} & \multicolumn{10}{c}{Zero-shot}                       \\
Mahalanobis \citep{DBLP:conf/nips/LeeLLS18} &55.89	&		99.33	&59.94	&		99.41	&65.96	&		98.54&	64.23		&	98.46&	61.50	&98.94 \\
Energy \citep{DBLP:conf/nips/LiuWOL20} & 85.09 &	81.08 &	84.24 &	79.02 &	83.38 &	75.08 &	65.56 &	93.65 	&79.57 &	82.21  \\
 ZOC \citep{esmaeilpour2022zero} & 86.09&	87.30&	81.20	&81.51	&83.39	&73.06	&76.46	&98.90	&81.79	&85.19 \\
MCM \citep{mingdelving}                 & 94.59 & 32.20 & 92.25 & 38.80 & 90.31 & 46.20 & 86.12 & 58.50 & 90.82     & 43.93    \\
 CLIPN \citep{wang2023clipn}                 & 95.27 & 23.94 & 93.93 & 26.17 & \textcolor{blue}{\textbf{92.28}}& 33.45 & 90.93 & 40.83 & 93.10     & 31.10    \\ 
NegLabel \citep{jiang2024negative} &
  \textcolor{blue}{\textbf{99.49}} &
  \textcolor{blue}{\textbf{1.91}} &
  \textcolor{blue}{\textbf{95.49}} &
  \textcolor{blue}{\textbf{20.53}} &
  91.64 &
  {35.59} &
  {90.22} &
  {43.56} &
  \textcolor{blue}{\textbf{94.21}} &
  \textcolor{blue}{\textbf{25.40}} \\ \midrule[1.0pt]
\multicolumn{1}{l}{} & \multicolumn{10}{c}{One-shot}                       \\
CoOp \citep{zhou2022learning} &91.26	&		43.38	&91.95	&		38.53	&89.09	&		46.68&	87.83		&	50.64&	90.03	&44.80 \\
LoCoOp \citep{DBLP:journals/corr/abs-2306-01293} & 92.49 &	38.49 &	93.67 &	33.27 &	91.07 &	39.23 &	89.13 &	49.25 	&91.59 &	40.06  \\
ID-like \citep{bai2024id} & 97.35 &	14.57&	91.08	&44.02	&91.08	&41.74	&\textcolor{red}{\textbf{94.38}}	&\textcolor{red}{\textbf{26.77}}	&93.47	&31.78 \\
SCT \cite{yu2025self} & 95.70 &	19.16&94.58	&23.52	&91.23	&\textcolor{blue}{\textbf{32.81}}	&86.66	&48.87	&92.04	&31.09 \\

  RAP(Ours) &
  \textcolor{red}{\textbf{99.55}} &
  \textcolor{red}{\textbf{1.48}} &
  \textcolor{red}{\textbf{96.35}} &
  \textcolor{red}{\textbf{17.55}} &
  \textcolor{red}{\textbf{93.68}} &
  \textcolor{red}{\textbf{27.13}} &
  \textcolor{blue}{\textbf{94.09}} &
  \textcolor{blue}{\textbf{27.25}} &
  \textcolor{red}{\textbf{95.92}} &
  \textcolor{red}{\textbf{18.35}} \\ \midrule[1.0pt]
\multicolumn{1}{l}{} & \multicolumn{10}{c}{Four-shot}                       \\
CoOp \citep{zhou2022learning} &92.60	&		35.36	&92.27	&		37.06	&89.15	&		45.38&	89.68		&	43.74&	90.93	&40.39 \\
LoCoOp \citep{DBLP:journals/corr/abs-2306-01293} & 93.93 &	29.45 &	93.24 &	33.06 &	91.32 &	41.13 &	90.54 &	44.15 	&92.26 &	36.95  \\
ID-like \citep{bai2024id} & 98.19 &	8.98&	91.64	&42.03	&90.57	&44.00	&\textcolor{red}{\textbf{94.32}}	&\textcolor{red}{\textbf{25.27}}	&93.68	&30.07 \\
SCT \cite{yu2025self} & 97.03 &	13.88&	94.85	&22.13	&92.09	&\textcolor{blue}{\textbf{30.20}}	&87.96	&45.53	&92.98	&27.93 \\
   RAP(Ours) &
  \textcolor{red}{\textbf{99.53}} &
  \textcolor{red}{\textbf{1.65}} &
  \textcolor{red}{\textbf{96.34}} &
  \textcolor{red}{\textbf{17.40}} &
  \textcolor{red}{\textbf{93.54}} &
  \textcolor{red}{\textbf{27.98}} &
  \textcolor{blue}{\textbf{94.19}} &
  \textcolor{blue}{\textbf{26.63}} &
  \textcolor{red}{\textbf{95.90}} &
  \textcolor{red}{\textbf{18.42}} \\ 
\bottomrule[1.5pt]
\end{tabular}%

}
\end{table}

\subsection{Main Experimental Results and Analysis}
Table \ref{standardOOD} presents the main experimental results. Notably, our proposed method exhibits strong performance in the 1-shot setting, significantly outperforming methods that require the complete training data. Specifically, methods using the full training data achieve a maximum average AUROC of 91.54\% and a minimum FPR95 of 37.93\%, while methods based on vision-language pre-trained models in few-shot or zero-shot settings achieve a maximum average AUROC of 94.21\% and a minimum FPR95 of 25.40\%. In contrast, our method, using only one sample per class in the 1-shot setting, achieves an average AUROC of 95.92\% and an average FPR95 of 18.35\%, significantly surpassing both methods with full training data and existing methods using zero-shot or few-shot ID training data. This demonstrates that RAP effectively leverages the potential of vision-language pre-trained models for OOD detection. Methods such as ID-like \citep{bai2024id} and NegLabel \citep{jiang2024negative} show inferior performance due to the inaccuracy and insufficiency of the supervision signals or external knowledge they introduce. In contrast, our method improves prompt accuracy using the joint similarity maximization principle and updates prompts based on real OOD samples during the test phase, reducing the FPR95 by 7.05\% and increasing the AUROC by 1.71\%. These results indicate that our proposed RAP alleviates the lack of effective supervision signals in current methods using valuable outlier representations. By introducing structured textual words from external knowledge, the proposed method provides more accurate and effective textual semantic supervision for prompt learning.

\begin{figure}[t]
  \centering
  \includegraphics[width=\textwidth]{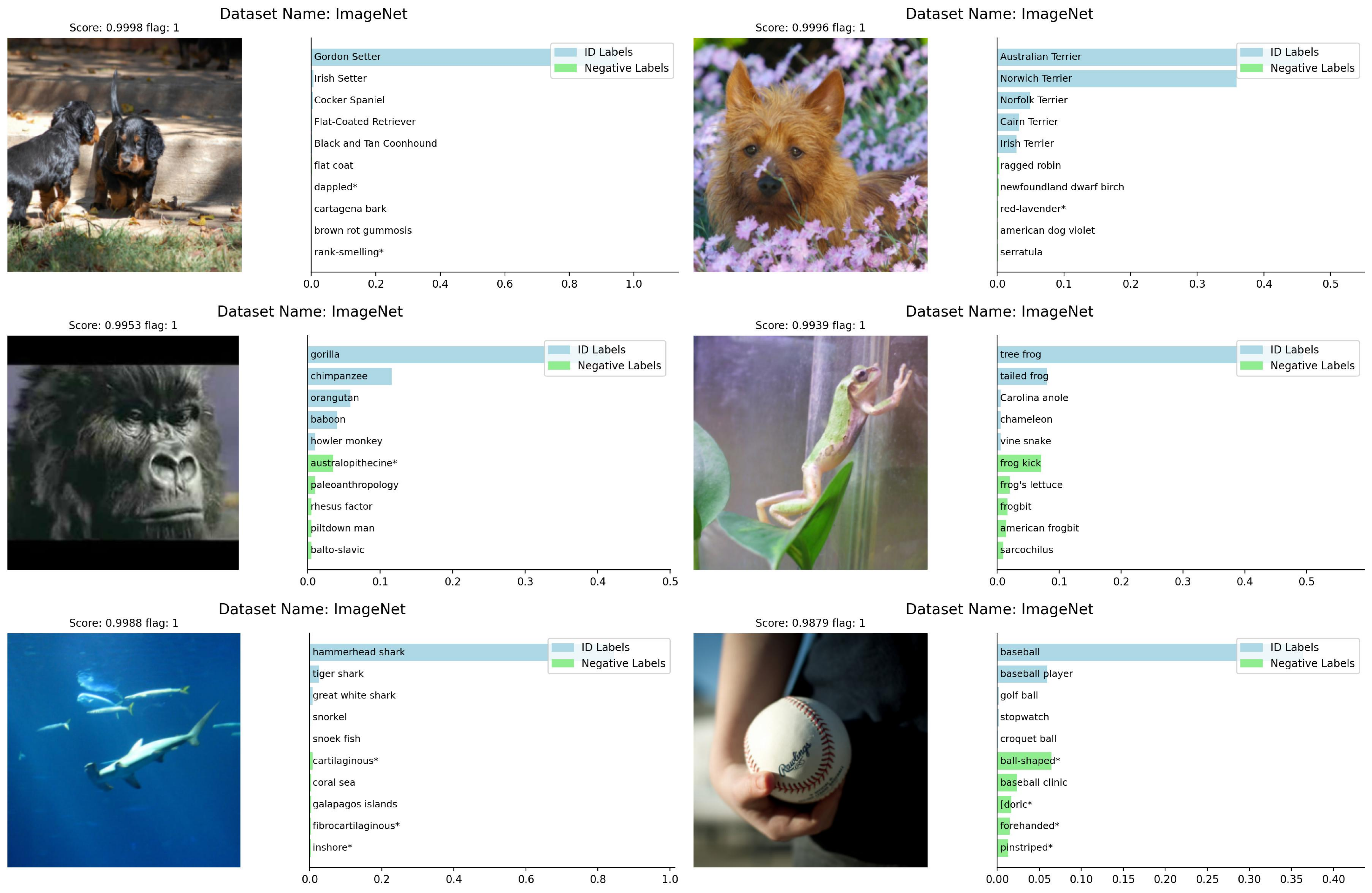}
  \caption{
    Case visualization of ID images. Each subfigure shows the original image on the left, along with its filename, dataset, OOD Detection score, and a flag indicating real ID (flag=1). The right side displays the softmax-normalized affinities to ID labels (blue) and negative labels (green) used by RAP.
  }
  \label{ID_case_study}
\end{figure}

\begin{figure}[]
  \centering
  \includegraphics[width=\textwidth]{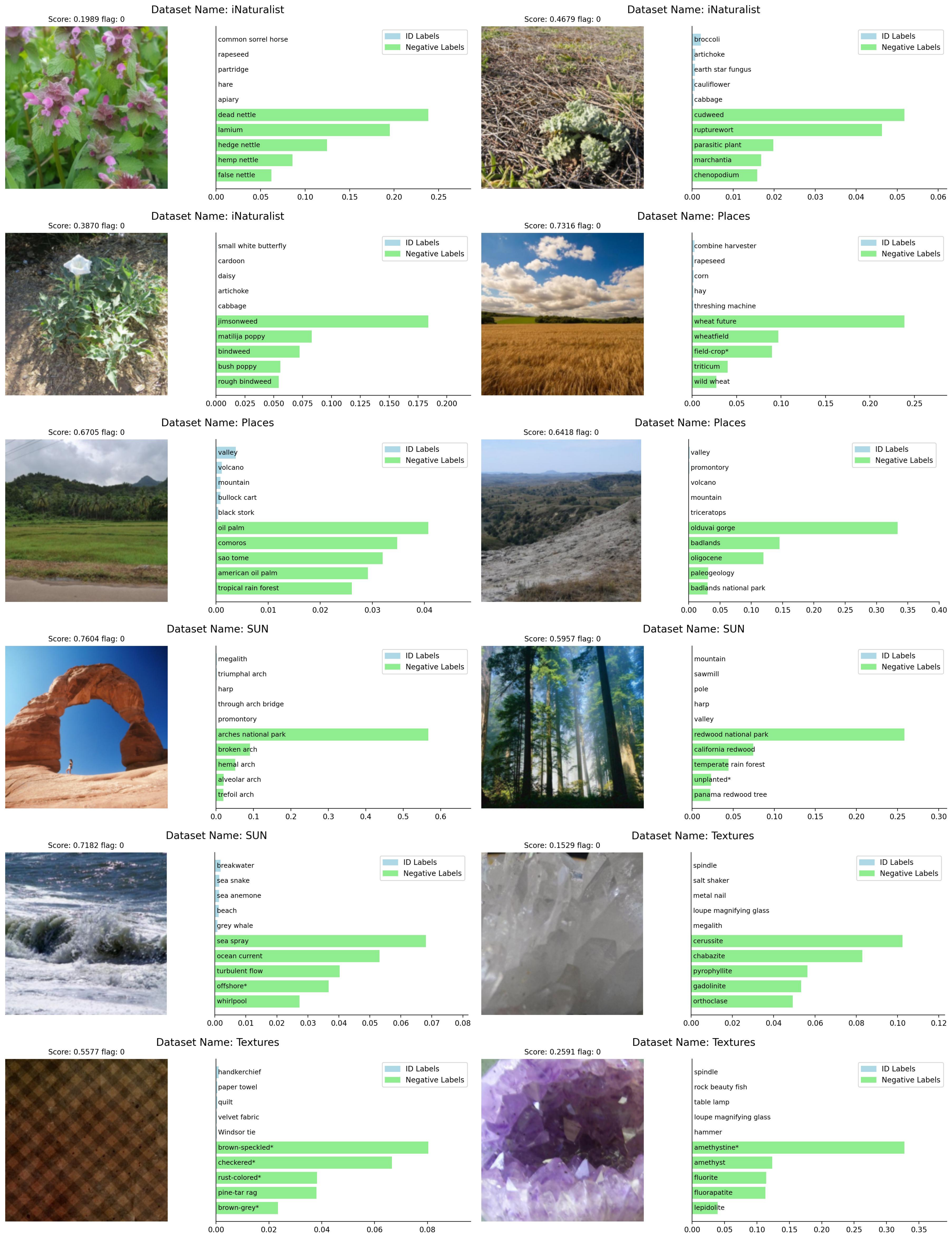}
  \caption{
    Case visualization of OOD images. Each subfigure shows the original image on the left, along with its filename, dataset, OOD Detection score, and a flag indicating real OOD (flag=0). The right side displays the softmax-normalized affinities to ID labels (blue) and negative labels (green) used by RAP.
  }
  \label{OOD_case_study}
\end{figure}

As shown in Figure \ref{ID_case_study} and Figure \ref{OOD_case_study} (zoom in for details), we conduct some case visualization analyses on the effectiveness of RAP on standard benchmarks. All images in Figure \ref{ID_case_study} are from the ID dataset ImageNet-1k \citep{huang2021importance}, while all images in Figure \ref{OOD_case_study} are from four OOD datasets: iNaturalist \citep{DBLP:conf/cvpr/HornASCSSAPB18}, Places \citep{DBLP:journals/pami/ZhouLKO018}, SUN \citep{DBLP:conf/cvpr/XiaoHEOT10}, and Textures \citep{DBLP:conf/cvpr/CimpoiMKMV14}, with three images selected from each dataset. Each subfigure consists of two parts: the left part displays the original image, its filename, dataset name, OOD detection score, and a flag indicating whether the sample is a real ID/OOD sample (flag=1 denotes a real ID instance, flag=0 denotes a real OOD instance). The right part presents the top-5 softmax-normalized affinities with respect to the ID (blue) and negative (green) labels used by RAP. The ID (blue) and negative (green) labels are combined with templates to form RAP's ID and OOD prompts. The external knowledge introduced by RAP can effectively capture the characteristics of real OOD samples in the test set while maintaining low similarity to ID samples. For example, the test image in the first row and first column of Figure\ref{OOD_case_study} is accurately described by the negative (green) label ``dead nettle'' and the image in the first row and second column is matched with the label ``jimsonweed'' These visualization analyses demonstrate the effectiveness of RAP in retrieving OOD prompts using the joint similarity maximization principle during training and updating OOD prompts based on potential real OOD samples at test time.

\begin{figure}[t]
  \centering
  \includegraphics[width=\textwidth]{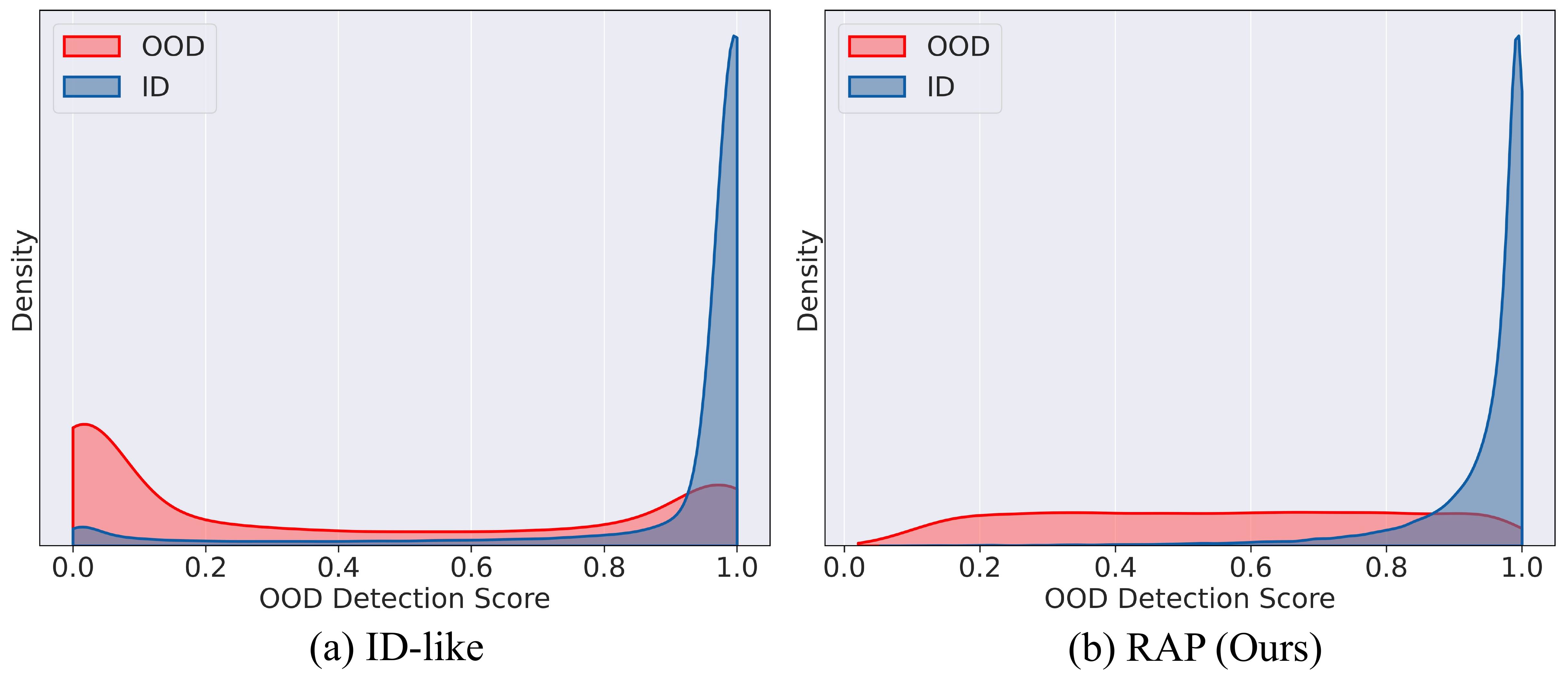}
  \caption{
    Density of the obtained ID and OOD scores with ID-like (left) and the proposed method RAP (right).
  }
  \label{fig_distribution}
\end{figure}

Figure \ref{fig_distribution} compares the score distributions of our method and ID-like \citep{bai2024id} on the ImageNet-1k benchmark with PLACES as the OOD dataset (zoom in for details). The x-axis represents the OOD Detection scores of the test samples, while the y-axis shows the probability distribution of these scores. The blue and red curves represent the score distributions of ID and OOD samples, respectively. Our method shows significantly less overlap between ID and OOD score distributions compared to ID-like \citep{bai2024id}. Specifically, ID-like \citep{bai2024id} retains a considerable number of ID samples with scores close to 0 and a large number of OOD samples with scores close to 1, indicating severe score distribution overlap. This makes it difficult to find a suitable threshold for distinguishing between ID and OOD data. In contrast, our method has nearly no ID samples with scores close to 0 and significantly fewer OOD samples with scores close to 1 compared to ID-like \citep{bai2024id}, demonstrating superior discriminative capability.

From Table \ref{standardOOD}, we observe that our method performs similarly in the 1-shot and 4-shot settings on the standard benchmark. This indicates that learning OOD representations with only one sample per class, combined with the additional accurate semantic supervision signals introduced by our method, is sufficient to achieve strong performance. However, on more challenging benchmarks, the 4-shot setting outperforms the 1-shot setting, suggesting that additional training samples are beneficial for datasets with higher semantic complexity, providing more comprehensive supervision information.

Furthermore, compared to gradient-based prompt learning methods that require hours of training, our proposed method takes only about 15 minutes during the training phase. During the test phase, when prompt updates are not required, RAP introduces only additional $O((P+Q)D)$ FLOPs per image (where \(P\) and \(Q\) are the numbers of OOD prompts for training and testing, respectively, and \(D\) is the embedding dimension), resulting in less than 1\% network latency. When prompt updates are required, RAP introduces additional $O(WD)$ FLOPs, where \(W\) is the number of WordNet words. The entire testing process takes only a few milliseconds per image, which highlights the efficiency of our method in real-time inference and rapid product iteration scenarios.

To further validate the performance of our method on difficult benchmarks with higher semantic similarity between ID and OOD classes, we follow the experimental setup of NegLabel and MCM, using ImageNet-10, ImageNet-20, and ImageNet-100 alternately as ID and OOD datasets to evaluate OOD detection performance. The results are shown in Table~\ref{hardOOD}, where the subscripts of the methods denote the number of samples per class used in training. For instance, ID-like$_1$ \citep{bai2024id} denotes one sample per class, and ID-like$_4$ denotes four samples per class.

\begin{table}[]
\normalsize
\centering
\caption{OOD detection performance comparison with multiple baselines on Challenging Benchmark. All values are percentages. The numerical subscripts in the method names denote the number of samples used per class. For example, RAP$_1$ corresponds to the 1-shot setting. Bold \textcolor{red}{\textbf{red}} indicates the best.}
\label{hardOOD}
\renewcommand{\arraystretch}{1.2}
\begin{tabular}{cc|c|cc}
\toprule[1.5pt]
\multicolumn{1}{c}{ID Dataset} & \multicolumn{1}{c|}{OOD Dataset} & Method & AUROC↑ & FPR95↓ \\ 
\midrule[1pt]

& & NegLabel \citep{jiang2024negative} & 98.58 & 5.40 \\
& & ID-like$_1$ \citep{bai2024id} & 92.52 & 46.30 \\
& & RAP$_1$(Ours) & \textcolor{red}{\textbf{98.98}} & \textcolor{red}{\textbf{3.20}} \\
& & ID-like$_4$ \citep{bai2024id} & 94.22 & 34.90 \\
\multirow{-4.8}{*}{ImageNet-10} & \multirow{-4.8}{*}{ImageNet-20} & RAP$_4$(Ours) & \textcolor{red}{\textbf{99.14}} & \textcolor{red}{\textbf{2.40}} \\
\midrule[1pt]
& & NegLabel \citep{jiang2024negative} & 97.33 & 14.00 \\
& & ID-like$_1$ \citep{bai2024id} & 92.86 & 42.80 \\
& & RAP$_1$(Ours) & \textcolor{red}{\textbf{98.23}} & \textcolor{red}{\textbf{7.20}} \\
& & ID-like$_4$ \citep{bai2024id} & 84.43 & 73.20 \\
\multirow{-4.8}{*}{ImageNet-20} & \multirow{-4.8}{*}{ImageNet-10} & RAP$_4$(Ours) & \textcolor{red}{\textbf{98.31}} & \textcolor{red}{\textbf{6.40}} \\
\midrule[1pt]
& & NegLabel \citep{jiang2024negative} & 85.40 & 64.00 \\
& & ID-like$_1$ \citep{bai2024id} & 81.76 & 82.20 \\
& & RAP$_1$(Ours) & \textcolor{red}{\textbf{88.40}} & \textcolor{red}{\textbf{53.60}} \\
& & ID-like$_4$ \citep{bai2024id} & 84.75 & 69.80 \\
\multirow{-4.8}{*}{ImageNet-100} & \multirow{-4.8}{*}{ImageNet-10} & RAP$_4$(Ours) & \textcolor{red}{\textbf{88.60}} & \textcolor{red}{\textbf{51.20}} \\
\bottomrule[1.5pt]
\end{tabular}
\end{table}

The results on the challenging benchmark further highlight the limitations of existing gradient-based few-shot methods. For example, ID-like \citep{bai2024id} struggles with difficult samples due to the scarcity of valuable outlier representations and the distributional differences between these representations and real test OOD samples, leading to overfitting and limited OOD detection performance. In contrast, our method introduces additional accurate and sufficient textual supervision signals from external knowledge for prompt learning, achieving superior performance.

\subsection{Ablation Study}
To validate the effectiveness of RAP during both the training and testing phases, we conducted ablation experiments on both standard and challenging benchmarks. Specifically, we performed the prompt retrieval augmentation as usual during the training phase, while in the testing phase, we did not apply the RAP (Test-time) and only used the ID prompts and the OOD prompts retrieved during training for OOD detection.

As shown in Table \ref{standardAblation} and Table \ref{hardAblation}, it is encouraging that even without RAP in the testing phase, the proposed method achieved an average AUROC of 94.71\% and an average FPR95 of 23.03\% on the standard benchmark. In contrast, ID-like \citep{bai2024id}, using one sample per class, only achieved an average AUROC of 93.49\% and an average FPR95 of 31.77\%, which is significantly lower than the performance of our proposed method.
\begin{table}[t]
\centering
\caption{Ablation study on whether to apply prompt retrieval augmentation during the training/testing phases on standard benchmarks.}
\label{standardAblation}
\resizebox{\textwidth}{!}{%
\begin{tabular}{@{}c|cccccccccc@{}}
\toprule[1.5pt]
                     & \multicolumn{8}{c}{OOD Dataset}                               & \multicolumn{2}{c}{} \\
 &
  \multicolumn{2}{c}{iNaturalist} &
  \multicolumn{2}{c}{SUN} &
  \multicolumn{2}{c}{Places} &
  \multicolumn{2}{c}{Textures} &
  \multicolumn{2}{c}{Average} \\ \cmidrule(l){2-3} \cmidrule(l){4-5}\cmidrule(l){6-7}  \cmidrule(l){8-9}\cmidrule(l){10-11}
\multirow{-3}{*}{Methods} &
  AUROC↑ &
  FPR95↓ &
  AUROC↑ &
  FPR95↓ &
  AUROC↑ &
  FPR95↓ &
  AUROC↑ &
  FPR95↓ &
  AUROC↑ &
  FPR95↓ \\  \midrule[1.0pt]

train &
  99.48 &
  1.78 &
  95.69 &
  20.16 &
  92.06 &
  34.01 &
  91.63 &
  36.19 &
  94.71 &
  23.03 \\  \midrule[1.0pt]
  
train\&test &
  \textbf{99.55} &
  \textbf{1.48} &
  \textbf{96.35} &
  \textbf{17.55} &
  \textbf{96.38} &
  \textbf{27.13} &
  \textbf{94.09} &
  \textbf{27.25} &
  \textbf{95.92} &
  \textbf{18.35} \\ 
\bottomrule[1.5pt]
\end{tabular}%

}
\end{table}

On the challenging benchmark, the improvement is even more significant. For instance, when using ImageNet-10 as the ID dataset and ImageNet-20 as the OOD dataset, our method achieved an average AUROC of 98.60\% and an average FPR95 of 4.60\%, significantly surpassing ID-like's AUROC of 92.52\% and FPR95 of 46.30\%. This highlights the advantage of using external knowledge to generate more effective prompts. Existing gradient-based prompt learning methods struggle with the scarcity of valuable OOD representations, resulting in insufficient supervision signals. In contrast, our method utilizes the joint similarity maximization principle to retrieve appropriate knowledge from external sources, providing accurate semantic supervision. These retrieved texts are directly used as OOD prompts for further prompt retrieval augmentation, leading to significant performance gains in OOD detection.







\begin{table}[]
\normalsize
\centering
\caption{Ablation study on whether to apply prompt retrieval augmentation during the training/testing phases on challenging benchmarks.}
\label{hardAblation}
\renewcommand{\arraystretch}{1.2}
\begin{tabular}{cc|c|cc}
\toprule[1.5pt]
\multicolumn{1}{c}{ID Dataset} & \multicolumn{1}{c|}{OOD Dataset} & Method & AUROC↑ & FPR95↓ \\ 
\midrule[1pt]
& & train & 98.60 & 4.60 \\
\multirow{-2}{*}{ImageNet-10} & \multirow{-2}{*}{ImageNet-20} & train\&test & \textbf{98.98} & \textbf{3.20} \\
\midrule[1pt]
& & train & 98.05 & 8.80 \\
\multirow{-2}{*}{ImageNet-20} & \multirow{-2}{*}{ImageNet-10} & train\&test & \textbf{98.23} & \textbf{7.20} \\
\midrule[1pt]
& & train & 87.66 & 57.60 \\
\multirow{-2}{*}{ImageNet-100} & \multirow{-2}{*}{ImageNet-10} & train\&test & \textbf{88.40} & \textbf{53.60} \\
\bottomrule[1.5pt]
\end{tabular}
\end{table}







\begin{table}[t]
\centering
\caption{Ablation study on each similarity in the joint similarity during prompt retrieval augmentation in the training phase on standard and challenging benchmarks.}
\label{DAblation}
\renewcommand{\arraystretch}{1.2}
\begin{tabular}{c|cccc}
\toprule[1.5pt]
 & \multicolumn{2}{c}{Standard Benchmark} & \multicolumn{2}{c}{Challenging Benchmark} \\ 
\cmidrule(l){2-3} \cmidrule(l){4-5}
\multirow{-2}{*}{Joint Similarity} & AUROC↑ & FPR95↓ & AUROC↑ & FPR95↓ \\
\midrule[1pt]
w.o. $sim_1$ & 94.95 & 21.27 & 93.20 & 32.40 \\
\midrule[1pt]
w.o. $sim_2$ & 94.66 & 23.38 & \textbf{95.23} & 21.70 \\
\midrule[1pt]
w.o. $sim_3$ & 95.55 & 19.97 & 94.73 & 25.87 \\
\midrule[1pt]
w. $sim_{1,2,3}$ & \textbf{95.92} & \textbf{18.35} & 95.20 & \textbf{21.33} \\
\bottomrule[1.5pt]
\end{tabular}
\end{table}

Furthermore, applying RAP during the testing phase further improves OOD detection. Since the OOD samples in the test set often differ from the valuable OOD representations used during training, existing methods face difficulties in distinguishing between ID and OOD data effectively. Our method mitigates this problem by selecting valuable OOD samples based on their OOD detection scores and retrieving matching words from external knowledge to update the OOD prompts. This adaptive retrieval strategy enhances the detection performance, demonstrating the robustness of the proposed method in test scenarios.

\begin{figure}[H]
    \begin{center}
    \includegraphics[width=1.0\textwidth]{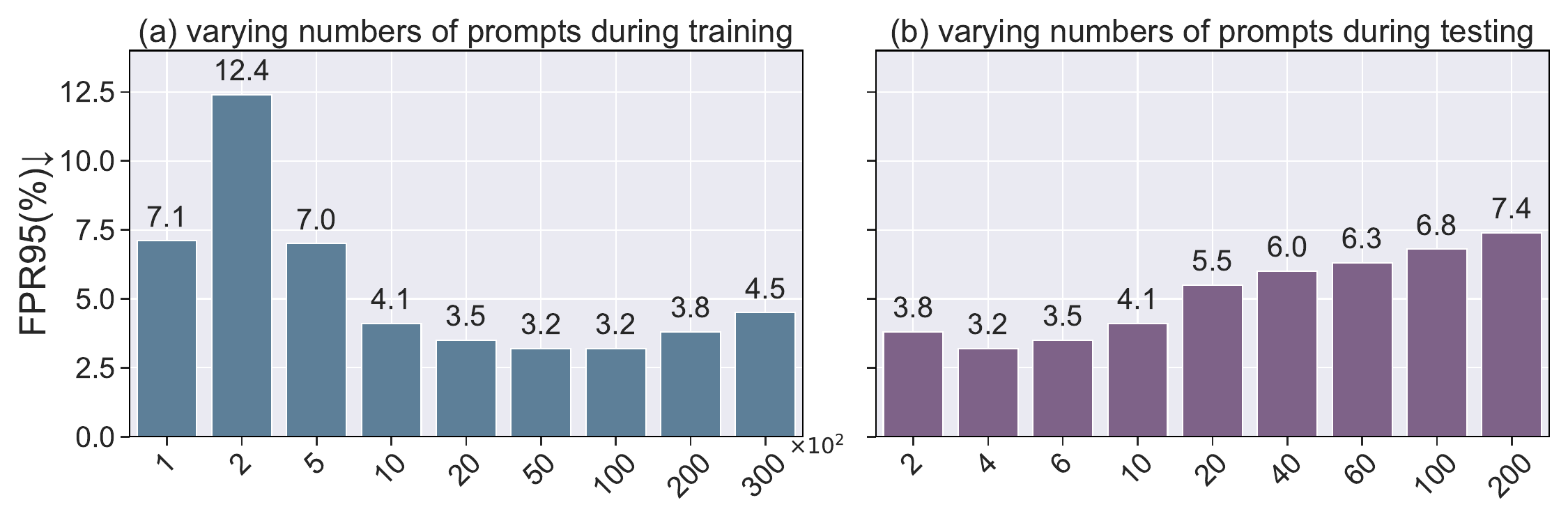}
    \end{center}
    \vspace{-15pt}
    \caption{
    Ablation study on the impact of the varying prompt numbers during both the training and testing phases on detection performance.
    }
    \label{prompts_num_traintest}
\end{figure}

To evaluate the contribution of each similarity in the joint similarity maximization principle, we conducted an ablation study by removing one similarity term at a time (\(sim_1\), \(sim_2\), or \(sim_3\)) and using the remaining two for RAP. The results on both standard and challenging benchmarks are shown in Table \ref{DAblation}. Specifically, on the standard benchmark, removing \(sim_1\), \(sim_2\), and \(sim_3\) led to an increase in FPR95 to 21.27\%, 23.38\%, and 19.97\%, respectively, and a decrease in AUROC to 94.95\%, 94.66\%, and 95.55\%. In comparison, using all three similarities resulted in an FPR95 of 18.35\% and an AUROC of 95.92\%. Similar trends were observed on the challenging benchmark, achieving an FPR95 of 21.40\% and an AUROC of 95.20\%, highlighting the importance of each similarity.

\(sim_1\) ensures that the OOD prompts are more aligned with OOD representations at a fine-grained image level, reducing false positives. Fine-grained image features, such as textures, edges, and background details, contribute to the accurate identification of OOD samples. \(sim_2\) and \(sim_3\) ensure sufficient separation between OOD prompts and ID representations in both abstract semantic and fine-grained image levels, minimizing distribution overlap.  
Each similarity contributes to optimizing the detection strategy from different perspectives, making the joint similarity maximization principle an effective approach for accurate text-based supervision.

To analyze the effect of the number of prompts used in RAP on detection performance, we evaluated the model with varying numbers of prompts during both the training and testing phases. The results are shown in Figure \ref{prompts_num_traintest}. When the number of retrieved prompts is insufficient, the model lacks the necessary semantic supervision signals, resulting in poor detection performance. This is particularly evident in complex OOD samples, where FPR95 increases significantly due to the model's reduced discriminative ability.

However, when too many prompts are retrieved, the model acquires excessive information, including irrelevant words with low relevance to the OOD samples. This introduces semantic noise into the OOD prompts, increasing the overlap between OOD and ID prompts and degrading the detection performance. Therefore, selecting an appropriate number of prompts is crucial for effective OOD detection. A balanced number of retrieved prompts ensures accurate OOD semantic supervision representation while minimizing the overlap between ID and OOD, maximizing the detection performance.

\section{Conclusion}
In this work, we propose a novel retrieval-augmented prompt learning framework (RAP) for OOD detection. By leveraging the rich textual information from external knowledge sources, we retrieve semantically relevant texts to enhance OOD prompts, thereby introducing effective additional semantic supervision. To address two fundamental challenges in current OOD detection methods—namely, the lack of usable supervision and the performance degradation caused by distribution shifts between training and testing data—we design Retrieval-Augmented Prompts During Training and Retrieval-Augmented Prompts During Testing. These components help bridge the gap between training and inference scenarios and improve the model's robustness to distributional discrepancies. Extensive experiments on multiple benchmarks, as well as case studies on real-world datasets, demonstrate the superior performance of our method in OOD detection. We further provide a comprehensive analysis of the efficiency of RAP and perform ablation studies that validate the effectiveness of each component, including the joint similarity strategy used for retrieval. Researchers in the field may deploy our method in real-world open-world testing environments to achieve fast and effective OOD detection. Moreover, our framework provides a flexible foundation for other researchers to incorporate more accurate and diverse external knowledge sources to further enhance detection performance. While our approach achieves strong OOD detection performance, it currently relies on external knowledge retrieved from large-scale text corpora only. This reliance may limit further performance gains, especially in more challenging OOD scenarios. As future work, we plan to explore integrating the rich and structured knowledge embedded in large language models (LLMs), which we believe could further enhance the semantic understanding and generalization capabilities of existing vision-language models for OOD detection.

\bibliographystyle{elsarticle-num} 
\bibliography{reference}

\end{document}